%% file: main.tex
\documentclass[sigconf]{acmart}
\input{packages}
\begin{document}
\title{Why Stop at Words? \\ Unveiling the Bigger Picture through Line-Level OCR}
\author{Shashank Vempati$^{1*}$ \quad Nishit Anand$^{2*}$ \quad Gaurav Talebailkar$^{3*}$ \quad Arpan Garai$^{4*}$ \quad Chetan Arora$^{5}$}
\authornote{All authors were at IIT Delhi while working on this project. Shashank and Nishit contributed equally.}
\affiliation{%
   \institution{%
   $^1$Typeface, India \quad 
   $^2$University of Maryland, College Park, USA \quad 
   $^3$Tata 1mg, India \quad \\
   $^4$Vellore Institute of Technology, Vellore, India \quad 
   $^5$Indian Institute of Technology Delhi, India
   }
   \city{\url{https://nishitanand.github.io/line-level-ocr-website}}
   \country{}
 }
\input{abstract}
\maketitle
\renewcommand{\shortauthors}{Vempati and Anand}
\vspace{-2pt}
\input{1_introduction}
\input{2_related_work}
\vspace{-4pt}
\input{3_proposed_dataset}
\input{4_proposed_approach}
\input{5_experiments_results}
\input{6_ablations_discussion}
\input{7_conclusion}
\input{8_future_work}
\bibliographystyle{ACM-Reference-Format}
\bibliography{bibliography}
\newpage
\appendix
\input{appendix}
\end{document}

%% file: packages.tex
\usepackage{booktabs} 
\usepackage{xspace}
\usepackage{float}                   
\usepackage{algorithm}               
\usepackage[noend]{algpseudocode}    
\usepackage{xcolor}                  
\setcopyright{rightsretained}
\newcommand{\etal}{{et al.}\xspace}
\acmYear{2025}
\copyrightyear{2025}
\acmConference[Line Level OCR]{Line Level OCR Workshop}{August}{2025}

\renewcommand\footnotetextcopyrightpermission[1]{} 

%% file: abstract.tex
\begin{abstract}
Conventional optical character recognition (OCR) techniques segmented each character and then recognized. This made them prone to error in character segmentation, and devoid of context to exploit language models. Advances in sequence to sequence translation in last decade led to modern techniques first detecting words and then inputting one word at a time to a model to directly output full words as sequence of characters. This allowed better utilization of language models and bypass error-prone character segmentation step. We observe that the above transition in style has moved the bottleneck in accuracy to word segmentation. Hence, in this paper, we propose a natural and logical progression from word level OCR to line-level OCR. The proposal allows to bypass errors in word detection, and provides larger sentence context for better utilization of language models. We show that the proposed technique not only improves the accuracy but also efficiency of OCR. Despite our thorough literature survey, we did not find any public dataset to train and benchmark such shift from word to line-level OCR. Hence, we also contribute a meticulously curated dataset of 251 English page images with line-level annotations. Our experimentation revealed a notable end-to-end accuracy improvement of $5.4\%$, underscoring the potential benefits of transitioning towards line-level OCR, especially for document images. We also report a $4$ times improvement in efficiency compared to word-based pipelines. With continuous improvements in large language models, our methodology also holds potential to exploit such advances. Full source code, models, and the annotated dataset will be released upon acceptance. 
\end{abstract}

%% file: 1_introduction.tex
\section{Introduction}
\begin{figure*}[t]
    \centering
    \begin{center}
    $\begin{array}{@{\hspace{1pt}}c}
    \includegraphics[height=2.52in]{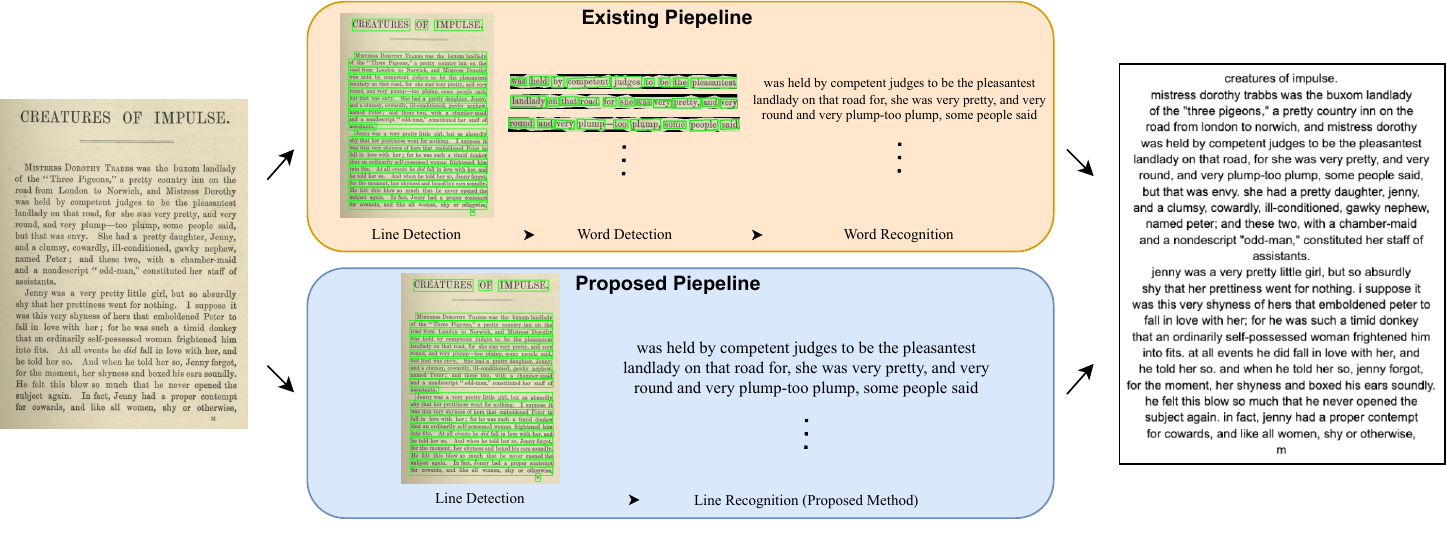}
    \end{array}$
    \end{center}
    \caption{Comparison between existing and proposed pipelines for OCR.}
    \label{fig:pipeline}
\end{figure*}
\subsection{Background}
In today's digital world, the ability to seamlessly convert physical documents into editable, searchable text is essential. Optical Character Recognition (OCR) technology bridges this gap, empowering individuals and businesses to unlock information trapped in hard copies. As the demand for advanced OCR solutions grows, organizations and researchers are actively developing systems that improve accuracy and efficiency to cater to diverse real-world needs. 
\subsection{Current OCR Pipeline}
End-to-end OCR, especially when working with entire documents, requires a carefully designed pipeline of multiple models, with text recognition as a crucial component. Most existing OCR systems employ a sequential pipeline: first detecting lines within the input page, then detecting words in each line. Finally, a recognition model identifies individual words. However, both line and word detection techniques are prone to errors. These errors propagate to the recognition model, compounding the overall error rate. Therefore, in this work, we propose a paradigm shift towards line-level recognition models to bypass the error-prone word detection step.
\vspace{-2pt}
\subsection{Limitations of Current OCR Techniques}
To understand OCR advancements, it is important to note that traditional OCR models broadly fall into two categories: character-level \cite{Drobac2020} and word-level \cite{bautista2022_parseq, fang2021ABINet, liu2016starNet}. Regardless of the chosen approach, these methods often require further processing to align with downstream applications that demand page-level outputs. Despite integration with additional models for end-to-end processing, these pipelines typically rely on sequential multi-stage segmentation (paragraph \cite{Moysset2015}, line \cite{dutta2021}, word \cite{Chen2022}, character \cite{Stewart2017}) before recognition. Unfortunately, even the most advanced layout parsers introduce errors at each stage, propagating and negatively impacting final accuracy. Moreover, skewed or warped pages disrupt word ordering in word-level OCR, causing accuracy errors. Hence, in this paper, we propose line-level OCR to minimize such problems.
\vspace{-2pt}
\subsection{Proposed Idea}
Recent advancements in computer vision have highlighted the efficacy of sequence-to-sequence models, particularly their ability to utilize contextual information to decipher even distorted characters. However, current OCR infrastructure primarily applies this capability at the word level. We propose a paradigm shift: extending these context-aware models to the line level. This approach can potentially overcome the limitations of existing word-centric methods, enabling them to tackle distorted parts, or highly ambiguous parts of the text within a sentence, and ultimately achieve superior recognition accuracy at the line level. For example, as we show in the experiments, our technique appears highly accurate for punctuation marks, which are very difficult to recognize at a word level, but becomes easy when seen in the context of a sentence.
\vspace{-2pt}
\subsection{Contributions}
This framework specifically targets the following challenges:
\begin{enumerate}
    \item Current end-to-end OCR systems often struggle with accuracy, especially when utilizing word or character segmentation methods. These methods can perform poorly on documents with various distortions. We propose to apply OCR directly at the line level, mitigating this error propagation. We report an accuracy improvement of $5.4\%$ using our proposed model compared to current state of the art of $92.15\%$ for CRAFT \cite{Baek_2019_CVPR_CRFT} based word detection, and PARSeq \cite{bautista2022_parseq} for word recognition.
    \item In documents with non-standard layouts (e.g., multiple columns, irregular text arrangements), identifying and isolating individual words becomes difficult. Segmentation algorithms tend to make mistakes, especially when words are adjacent or if spaces are inconsistent. By minimizing reliance on accurate word segmentation, the proposed framework tackles a significant issue in traditional OCR. This approach highlights why line-level OCR is essential for handling complex and varied document layouts.
    \item As the additional word-detection step is waived from the pipeline, the overall end-to-end time efficiency is also improved. We report a four-fold efficiency improvement using our proposed model compared to current state of the art.
    \item To evaluate the accuracy of various models in an end-to-end manner (including detection and recognition), we need a dataset of images at the page level. To the best of our knowledge, there are no such datasets publicly available. Hence, to fill this gap, we contribute a dataset of 251 images annotated at the page-level. 
\end{enumerate}
\begin{figure*}
  \centering
  \setlength{\tabcolsep}{2pt}        
  \renewcommand{\arraystretch}{0.8}   
  \begin{tabular}{*{9}{c}}
    \includegraphics[height=1.3in]{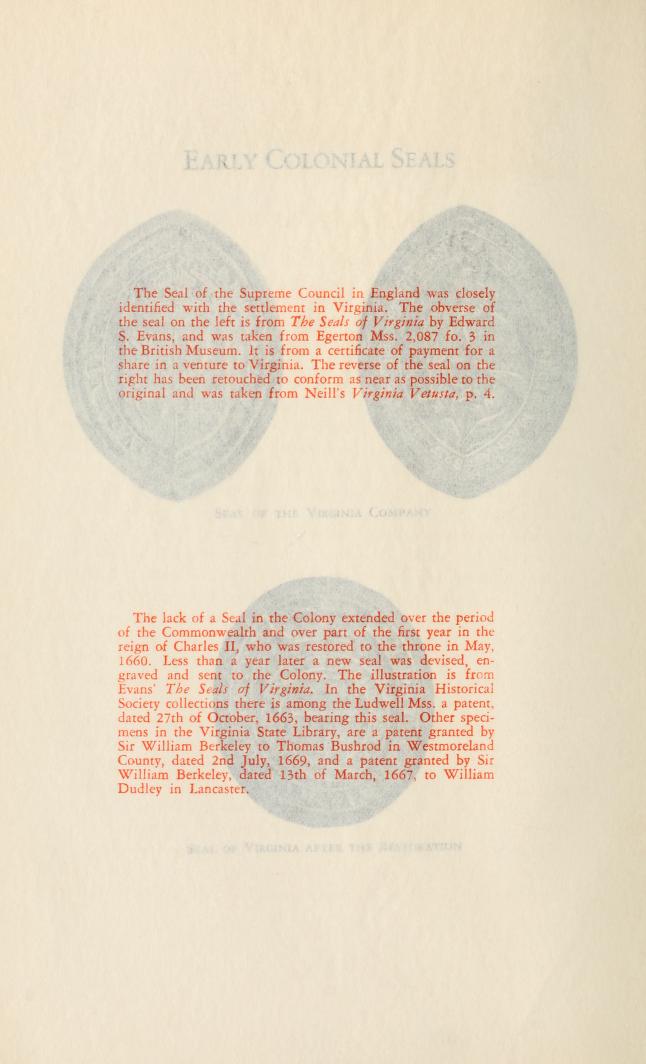} &
    \includegraphics[height=1.3in]{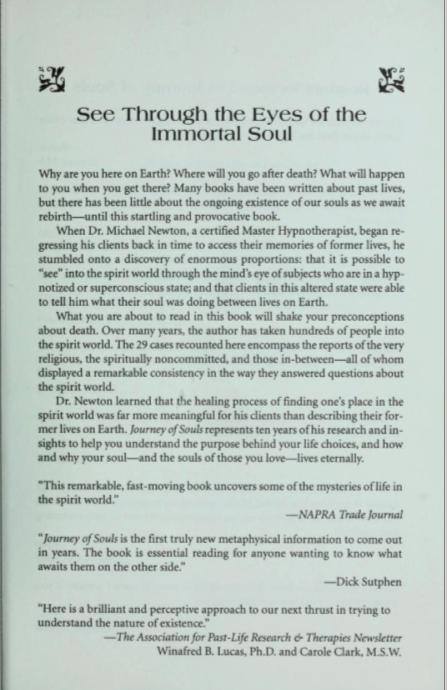} &
    \includegraphics[height=1.3in]{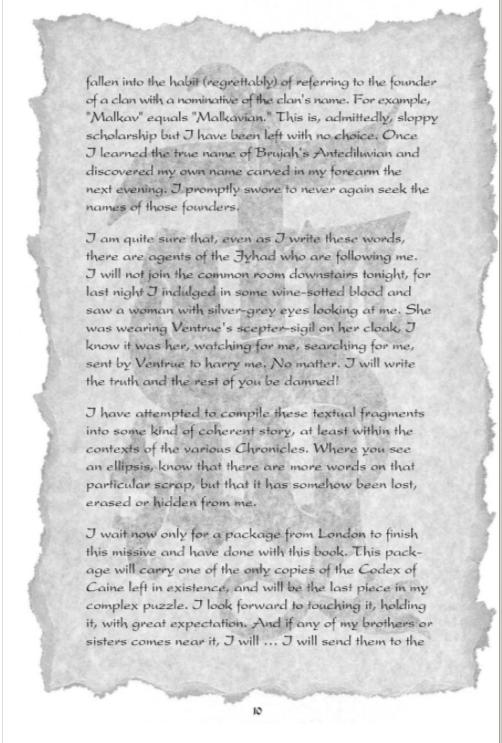} &
    \includegraphics[height=1.3in]{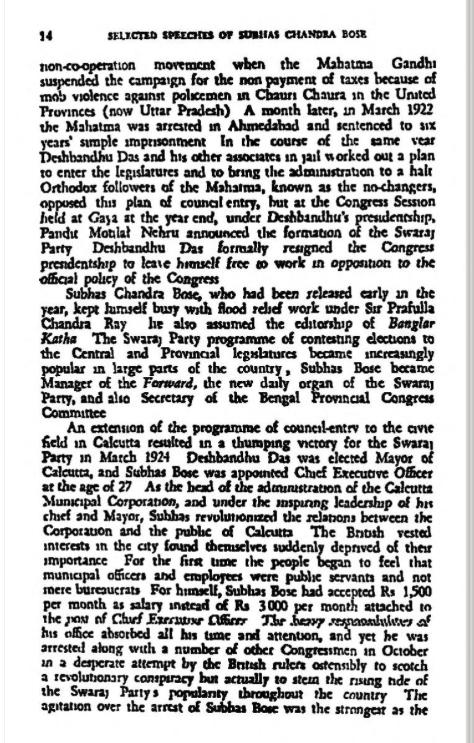} &
    \includegraphics[height=1.3in]{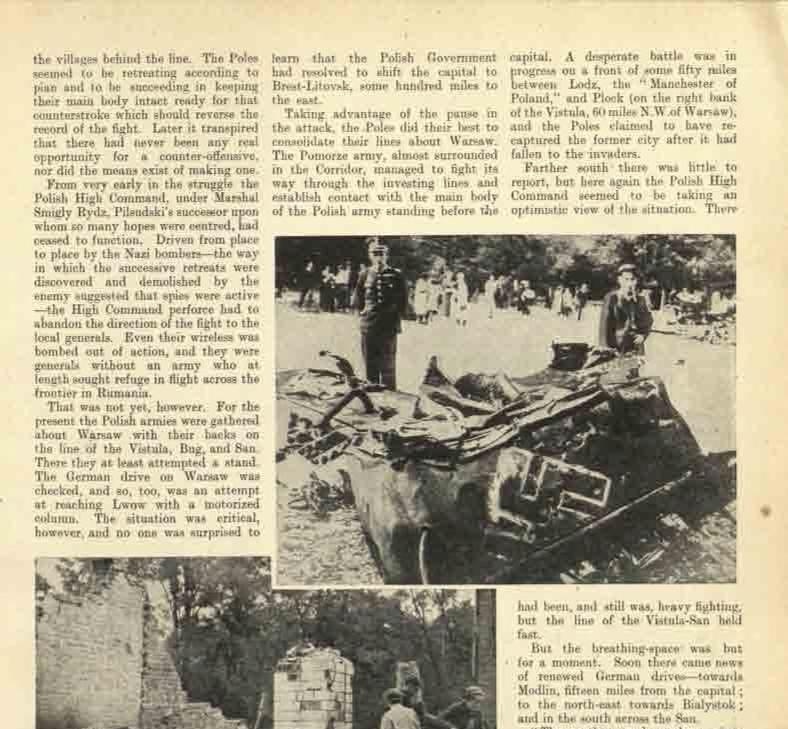} &
    \includegraphics[height=1.3in]{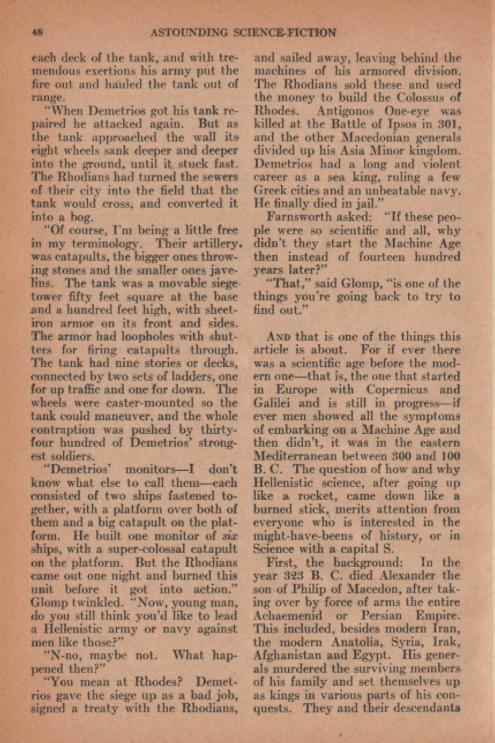} &
    \includegraphics[height=1.3in]{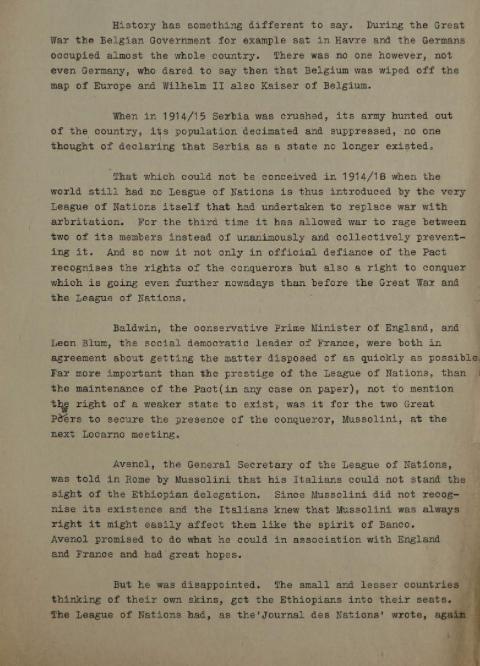}
  \end{tabular}
  \caption{Sample images from the proposed dataset. Additional images are included in the supplementary material.}
  \label{fig:dataset}
\end{figure*}

%% file: 2_related_work.tex
\section{Related Work}
\subsection{Evolution of OCR Techniques}
Early OCR techniques relied on matching scanned characters to predefined templates. Later approaches involved feature engineering techniques like HOG (Histogram of Oriented Gradients) combined with SVMs (Support Vector Machines) for classification. The advent of Convolutional Neural Networks (CNNs) revolutionized OCR, as seen in models like LeNet-5 \cite{lecun1998gradient}. Recurrent Neural Networks (RNNs) and later LSTM architectures enabled sequential character recognition. Hybrid CRNN models further advanced sequence recognition accuracy.  Modern advancements integrate attention mechanisms popularized by the Transformer architectures \cite{vaswani2017attention}, leading to models like STAR-Net \cite{liu2016starNet} and ABINet \cite{fang2021ABINet}. In recent times, techniques like Permutation Language Modeling \cite{bautista2022_parseq} and Glyph-centric attention \cite{guan2023selfsiga,guan2023selfccd} have demonstrated notable advancements in the recognition aspect of OCR. 
\subsection{Role of Context and Language Models in OCR}
Modern text recognition systems typically begin by detecting regions of interest containing text within an image \cite{Chen2022,liao2022DBNETpp}. OCR models then analyze these regions using varying strategies. Context-free OCR models treat characters as isolated units, predicting them without considering the broader text \cite{shi2016,liu2016starNet}. Context-aware OCR models utilize semantics and linguistic patterns to improve recognition accuracy, often employing recurrent neural networks (RNNs) \cite{bahdanau2014neural} or transformers \cite{sheng2019nrtr,lee2020recognizing,bleeker2019bidirectional}. Specific models have extended their capabilities by using external language models \cite{mansimov2019generalized,fang2021ABINet}, leveraging their understanding of typical language structure to inform character-level predictions. Language models are incorporated using diverse techniques, such as two-stream attention-based masked language modeling (MLM) \cite{qi2021bang}, permutation language modeling (PLM) \cite{bautista2022_parseq,tian2020train}, and two-stream attention parameterization \cite{yang2019xlnet} techniques.
\subsection{Word Detection Techniques}
Text detection is a crucial precursor step within the OCR pipelines. Prior to recognizing characters and forming words, the system must isolate the areas of the image that actually contain text. This becomes incredibly important in complex documents or natural scenes where text might be nestled among non-text elements, appear in irregular orientations, or exhibit variations in size and style. Examples of modern text detection models include Kraken \cite{kiessling2020modularkraken}, DBNet (DB) \cite{liao2020Dbnet,liao2022DBNETpp}, DPText-DETR (DPText) \cite{ye2023dptextdetr}, TextFuseNet (TFN) \cite{ye2020textfusenet}, CRAFT \cite{Baek_2019_CVPR_CRFT}, MixNet \cite{zeng2023mixnet}, SRFormer \cite{bu2023srformer} etc. These models use diverse techniques to manage challenges like complex layouts and irregular text orientations. One method involves creating a `probability map' to differentiate text from background areas. Another powerful approach leverages transformers, treating text detection as a direct set prediction problem. Additionally, integrating various levels of context and image features enhances the ability to handle complex scenarios containing diverse text appearances.
\vspace{4pt}
\subsection{End-to-end OCR Accuracy}
While recent advances in word-based OCR have yielded impressive results, particularly for printed and scene text recognition, reaching near-saturation levels of accuracy, challenges remain when considering document understanding at the page or line level. Word-based models often overlook subtle punctuation or small characters, especially within printed text, leading to errors that stem from both imperfections in detection algorithms and limitations of the word-level approach itself.  On a page-level, detection errors further increase, reducing overall line and page recognition accuracy compared to isolated word analysis. While  achievements in word-level OCR are laudable, a clear need exists for improving end-to-end accuracy at larger structural levels within documents.

%% file: 3_proposed_dataset.tex
\vspace{4pt}
\section{Proposed Dataset}
To facilitate research and benchmarking end-to-end optical character recognition (OCR) at the page-level, we introduce a new annotated dataset encompassing distinct linguistic, quality and layout challenges. 

We meticulously curated a dataset of 251 page images from archives\footnote{\url{https://archive.org}}, each selected to introduce a variety of recognition challenges commonly encountered in real-world OCR applications.  The images were sourced from diverse materials, including storybooks and printed versions of online documents, with publication dates spanning from 1862 to 2024. This temporal range ensures the dataset reflects the historical and stylistic variations found in printed text. 

To closely simulate real-world OCR difficulties, the dataset incorporates challenges such as long sentences, multi-column layouts, blurry or faded text, warped pages, dark backgrounds with light text, watermarks, embedded figures, and a wide range of fonts and typesetting styles.  This intentional diversity provides a robust test-bed for evaluating end-to-end OCR systems. Sample images are provided in Fig. \ref{fig:dataset}, while Fig. \ref{fig:stats} offers detailed dataset statistics.

%% file: 4_proposed_approach.tex
\section{Proposed Approach}
 \vspace{4pt}
\subsection{Line Detection} 
We employ Kraken \cite{kiessling2020modularkraken}, a line detection model known for its robustness and modular design. Kraken segments the input image into individual text lines. We can represent this step as $L = \text{Kraken}(I)$, where $I$ is the input RGB image, and $L = \{L_1, \ldots, L_m\}$ is the set of detected text lines. Following steps are followed to do the line detection: 
\begin{enumerate}
    \item \textbf{Multi-Label Pixel Classification:} In the first step we classify each pixel in the input image into three classes, belonging to baselines, text regions, or auxiliary classes (start/end separators). Kraken uses a deep neural network with downsampled encoder, and atrous convolutions to efficiently increase the receptive field, followed by unidimensional LSTM layers for context-aware analysis. This architecture reduces memory requirements compared to  conventional U-Net approaches. Readers are referred to \cite{kiessling2020modularkraken} for more details of the architecture.
    \item \textbf{Baseline Extraction and Polygonization:} In the second step, we extract baselines from the classification output. We then use seam-carving and the original image to compute accurate bounding polygons for each baseline.
    \item \textbf{Region Extraction:} In the last step, we use contour finding to extract text regions from their respective heatmaps. Notably, we allow baselines to exist independently, and to cross region boundaries for robustness.  
\end{enumerate}
\begin{figure*}
\includegraphics[width=\linewidth]{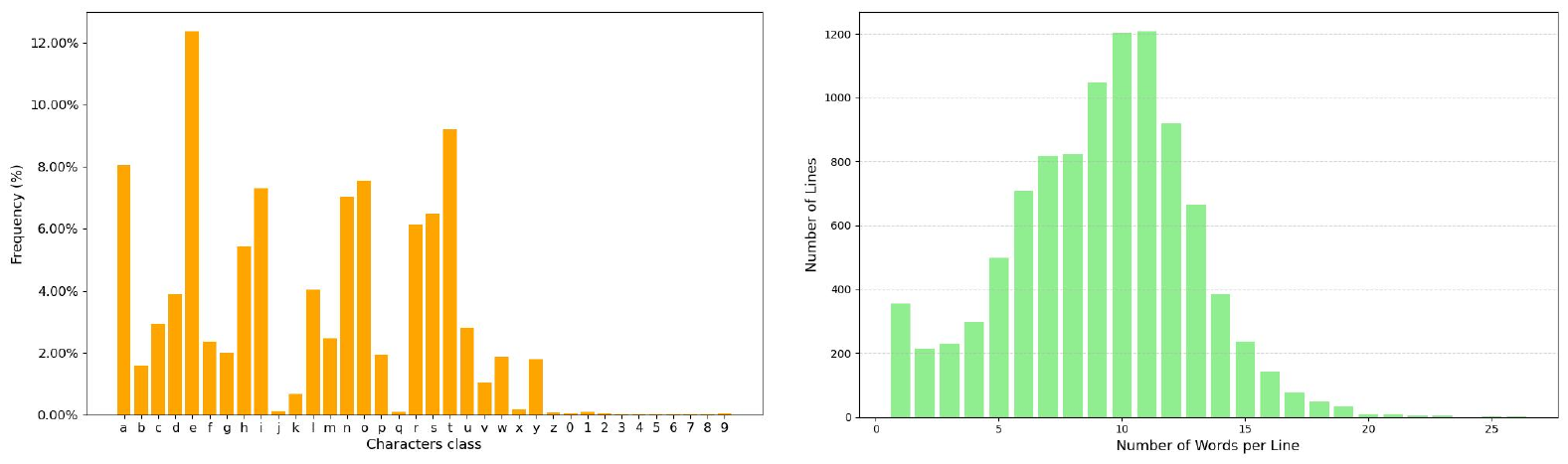}\\
\caption{Left: Character frequency distribution, and Right: Words per Line distribution in our proposed dataset.}
\label{fig:stats}
\end{figure*}
\begin{figure*}
    \centering
    \begin{center}
    $\begin{array}{@{\hspace{1pt}}c@{\hspace{1pt}}c@{\hspace{1pt}}c}
\includegraphics[height=0.32in]{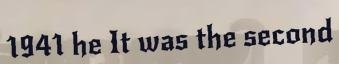}&
\includegraphics[height=0.32in]{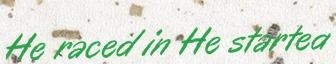}&
\includegraphics[height=0.32in]{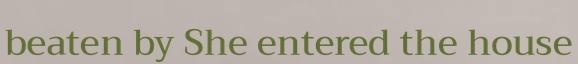}\\
\end{array}$
$\begin{array}{@{\hspace{1pt}}c@{\hspace{2pt}}c}
\includegraphics[height=0.32in]{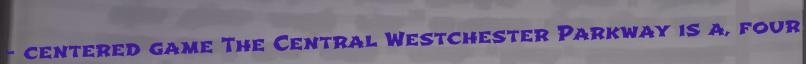}&
\includegraphics[height=0.32in]{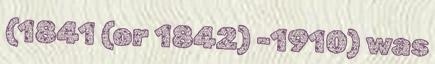}
 \end{array}$
	\end{center}
    \caption{Sample synthetic line-level images generated using TRDG}
    \label{fig:samples}
\end{figure*}
\subsection{Line-Level Recognition} 
We utilize PARSeq \cite{bautista2022_parseq} architecture, and fine-tune it for line-level recognition. PARSeq leverages permuted autoregressive language modeling and cross-attention mechanisms to effectively integrate visual and linguistic understanding. The recognition step can be denoted as a set of recognised text lines $T = \{T_1, \ldots, T_m\}$ where $T_i = \text{PARSeq}_\text{line}(L_i)$. The steps are as follows:
\begin{enumerate}
    \item \textbf{Line Image ($L$):} The input to the model is a preprocessed cropped image $L_i$ representing the detected text line.
    \item \textbf{Encoder-Decoder Architecture:} Encoder ($f_{enc}$) module takes the line image $L_i$ as input and extracts visual features. We can denote the encoded representation as $Z_i = f_{enc}(L_i)$. On the other hand, Decoder ($f_{dec}$) module decodes the encoded features $Z_i$ into the predicted text sequence $T_i$.
    \item \textbf{Permuted Autoregressive Sequence Modeling:} We use PARSeq to leverage the concept of permutation language modeling, as introduced in the XLNet model \cite{yang2019xlnet}. This approach deviates from traditional autoregressive models that predict text sequences strictly from left to right.  Instead, permutation language modeling randomly shuffles the order of characters during training, forcing the model to learn more robust bidirectional representations of the sequence. Important components of this module includes: 
    \begin{enumerate}
        \item Permutation Matrix ($\pi$):  The proposed pipeline employs a permutation matrix $\pi$. This matrix randomly shuffles the order of character positions within the line.
        \item Parallel Processing: During decoding, the model predicts all characters in the permuted order simultaneously, leveraging the power of parallel processing. This leads to faster inference speeds compared to strictly autoregressive approaches.
        \item Character Embedding ($E$):  The decoder utilizes a character embedding matrix $E$ to map each predicted character position to a corresponding embedding vector.
    \end{enumerate}
    \item \textbf{Attention Mechanism ($f_\text{attn}$):} A crucial aspect of proposed pipeline is the cross-attention mechanism ($f_{attn}$). This allows the decoder to focus on relevant parts of the encoded features $Z_i$ while predicting each character. We can denote the attention weights as \\$A = f_{attn}(Z_i, E(\Pi(T_i)))$.
    \item \textbf{Output Layer:} The final layer uses a softmax function to convert the decoder outputs into a probability distribution over the vocabulary as follows:
    \begin{equation}\label{par_seq_eq}
    \text{arg max}_{T_i} ~ \mathbb{P} \big( T_i | Z_i \big) \approx \text{arg max}_{T_i} ~ \prod ~ \mathbb{P} \big( t_{i,j} \mid Z_i, ~ T_{(i, \prec j)}, ~ \pi(T_i) \big) .
    \end{equation}
    Here, $i$ denotes the index of the line image and $j$ is the character that is being predicted in the $i^\text{th}$ line image. $T_i$ denotes the predicted text sequence, and $Z_i = f_\text{enc}(L_i)$ denotes the encoded representation corresponding to $i^\text{th}$ line image. Further, $t_{i,j}$ denotes the distribution of the $j^\text{th}$ character in the $i^\text{th}$ line, and $T_{(i, \prec j)}$ represents all characters in the $i^\text{th}$ line predicted before the $j^\text{th}$ position. $\mathbb{P}(t_{i,j} \mid \ldots)$ is the conditional probability of the character at position $j$ in line $i$, given the encoded features of that line, previously predicted characters, and the permuted order.
\end{enumerate}
\begin{table*}[t]
\begin{center}
\setlength{\tabcolsep}{18pt}
\resizebox{\linewidth}{!}{
\begin{tabular}{ ccccc }
\toprule
Word Detection & Word Recognition & CRR ($\uparrow$) & Flex Character Acc. ($\uparrow$) & Inference Time (in sec) ($\downarrow$)
\\
\midrule
DBNet & PARSeq & 81.94 & 89.66 & 05.64 \\
DBNet & ABINet & 80.72 & 88.48 & 05.91\\
DBNet & MATRN & 79.66 & 85.63 & 06.39\\
DBNet & CCD & 75.78 & 83.41 & 14.27\\
DBNet & MAERec & 81.26 & 89.26 & 16.20\\
DBNet & SIGA & 77.36 & 86.13 & 06.92\\
\midrule
DPText & PARSeq & 83.02 & 91.75 & 07.08\\
DPText & ABINet & 82.43 & 90.88 & 07.35\\
DPText & MATRN & 81.04 & 87.82 & 07.83\\
DPText & CCD & 79.69 & 88.23 & 15.71\\
DPText & MAERec & 82.75 & 91.48 & 17.64\\
DPText & SIGA & 79.66 & 88.76 & 08.45\\
\midrule
TextFuseNet & PARSeq & 80.48 & 89.57 & 29.29\\
TextFuseNet & ABINet & 80.02 & 89.05 & 29.57\\
TextFuseNet & MATRN & 78.80 & 86.24 & 30.05\\
TextFuseNet & CCD & 77.35 & 85.90 & 37.93\\
TextFuseNet & MAERec & 77.97 & 87.18 & 39.86\\
TextFuseNet & SIGA & 77.80 & 86.74 & 30.63\\
\midrule
CRAFT & PARSeq & \color{blue}{83.46} & \color{blue}{92.15} & \color{blue}{02.11}\\
CRAFT & ABINet & 82.68 & 91.30 & 02.39\\
CRAFT & MATRN & 81.03 & 87.49 & 02.88\\
CRAFT & CCD & 80.04 & 88.59 & 10.76\\
CRAFT & MAERec & 83.01 & 91.83 & 12.69\\
CRAFT & SIGA & 77.33 & 86.22 & 03.35\\
\midrule
MixNet & PARSeq & 82.52 & 91.25 & 10.38\\
MixNet & ABINet & 81.86 & 90.34 & 10.65\\
MixNet & MATRN & 80.41 & 87.08 & 11.13\\
MixNet & CCD & 75.67 & 83.10 & 19.01\\
MixNet & MAERec & 82.19 & 90.92 & 20.94\\
MixNet & SIGA & 79.25 & 88.39 & 11.72\\
\midrule
- & $\text{PARSeq}_\text{line}$ & \textbf{85.76} & \textbf{97.62} & \textbf{00.53} \\
\bottomrule
\end{tabular}
}
\end{center}
\caption{Pipelines comparison with various models at different stages on our {\em English} dataset. Kraken is commonly used as the line detection model for all experiments. Inference time is calculated per page excluding Kraken's time.}
\label{tab:WordAndLineComparison}
\end{table*}

%% file: 5_experiments_results.tex
\section{Experiments and Results}
As previously discussed, traditional OCR pipelines generally involve three stages: line detection, word detection, and word recognition. In contrast, we propose a framework that replaces the last two stages, word detection and recognition, with a single stage line recognition. We adopted the state-of-the-art Kraken line detection model \cite{kiessling2020modularkraken} for consistent line segmentation in our experiments as well as the proposed pipeline. For our investigation, we employed five state-of-the-art (SOTA) word detection models. Additionally, we retrained six SOTA word recognition models using fixed charsets and standard synthetic datasets. Line recognition was achieved using the PARSeq model, which was trained on synthetically generated line-level data. Please note that training recognition models only using synthetic data is inline with the trend in the recent SOTA \cite{Jiang_2023_ICCV_MAERec, guan2023selfsiga, wang2022multi_MGP_STR}. The following subsections provide details on the synthetic image generation process, experimental settings, and the evaluation metrics used.
\subsection{Data generation}
\vspace{8pt}
For line-level images, we utilize the Text Recognition Data Generator (TRDG\footnote{\url{https://github.com/Belval/TextRecognitionDataGenerator}}) \cite{trdg} for synthetic images generation. We've curated a diverse font collection of $3309$ English fonts to ensure realistic text rendering. Our English corpus comprises $1.21$ million text lines containing $600,337$ distinct words sourced from Wikipedia and other sources \cite{kakwani2020indicnlpsuite}. \\
Along with standard characters, we tried including commonly used special characters and numerals into our vocabulary to enhance its robustness. TRDG's flexibility enables us to simulate a variety of image distortions during synthetic image generation. We detail these in the supplementary material. \\
We generated a synthetic image dataset at the line-level for English, totaling $6$ million images.
For experiments involving word-level models, we relied on established synthetic datasets: SynthText \cite{Gupta16}, which comprises $5.5$ million images, and MJSynth \cite{Jaderberg16}, containing $8.9$ million images.
\subsection{Experimental setup}
For model training, we deployed a computational infrastructure with 32 CPUs and 8 NVIDIA A100 GPUs. To ensure comprehensive evaluation across diverse scenarios, models were tested against a newly created benchmark dataset.
\begin{figure*}[h]
    \centering
    \begin{center}
    $\begin{array}{@{\hspace{8pt}}c@{\hspace{8pt}}c}
\includegraphics[height=0.27in]{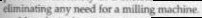}&
\includegraphics[height=0.27in]{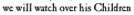}\\
\text{\scriptsize{GT: eliminating any need for a milling machine}} & \text{\scriptsize{   GT: we will watch over his children.}}\\
\text{\scriptsize{CRAFT+ABINet: eliminating any need.fora.miling.mahhiee}} & \text{\scriptsize{   DBNet+PARSeq: wewillwatch overhis children}}\\
\text{\scriptsize{PARSeq$_{line}$: eliminating any need for a milling machine,}} & \text{\scriptsize{   PARSeq$_{line}$: we will watch over his children}}\\
\end{array}$
\newline
$\begin{array}{@{\hspace{1pt}}c@{\hspace{8pt}}c}
\includegraphics[height=0.27in]{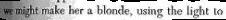}&
\includegraphics[height=0.27in]{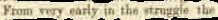}\\
\text{\scriptsize{GT: we might make her a blonde, using the light to}} & \text{\scriptsize{   GT: from very early in the struggle the}}\\
\text{\scriptsize{Mixnet+MATRN: weinight make her blonde using the light to}} & \text{\scriptsize{   CRAFT+CCD: from:}}\\
\text{\scriptsize{PARSeq$_{line}$: wemight make her a blonde, using the light to}} & \text{\scriptsize{   PARSeq$_{line}$: from very early in the struggle the}}
\end{array}$
	\end{center}
    \caption{Challenging examples for end-to-end word pipelines.}
    \label{fig:failure_word_recognition}
\end{figure*}
\subsection{Implementation Details}
We trained several word-level recognition models using a consistent character set: PARSeq \cite{bautista2022_parseq}, ABINet \cite{fang2021ABINet}, MATRN \cite{na2022multi}, SIGA \cite{guan2023selfsiga}, CCD \cite{guan2023selfccd}, and MAERec \cite{Jiang_2023_ICCV_MAERec}. Since ABINet and MATRN rely on external language models (LMs), we utilized the WikiText-103 dataset \cite{wikitext103} for LM training. \\
All word-models are trained on synthetic word images with a fixed image size of $32\times128$ pixels and utilized a $95$-character English charset. We extended our experimental setup for line recognition by training PARSeq on line-level images, while maintaining a fixed image size of $32\times400$ pixels. This involved modification of the charset to include a space character. Furthermore, we integrated the trained word-level models with both word and line detection models to facilitate end-to-end OCR evaluation. \\
Similarly, for the line-level approach, we exclusively employed our own proposed method using PARSeq, which was directly combined with a line detection model for comprehensive end-to-end evaluation of results.
\vspace{-4pt}
\subsection{Accuracy Metrics}
Character Recognition Rate (CRR) is a widely adopted metric for evaluating text recognition OCR models. We give details of the metric in supplementary material. \\ 
However, despite its widespread adoption, CRR suffers from several inherent limitations. One significant drawback lies in its heavy reliance on page and line segmentation methods, as well as the ordering of lines subsequent to the text line detection phase. Furthermore, CRR's inability to address contextual errors, such as misinterpretation of characters due to surrounding text or formatting inconsistencies, may hinder a comprehensive understanding of the model's end-to-end capabilities, leading to a significant decline in accuracy. \\
To address these shortcomings, we explored Flexible Character Accuracy (FCA), as introduced by Clausner \etal \cite{clausner2020flexible}. Unlike CRR, FCA is independent of line reading order, which is particularly advantageous given the challenge of predicting how a page of text will be read by humans, as well as the precise location of each character. By employing a hybrid computation that combines CRR at the line-level with the flexibility of page-level computations, FCA enables accurate measurement of order-independent accuracy, thereby enhancing the evaluation of OCR models.
\subsection{Quantitative Results}
Our quantitative results on an English page-level dataset (Table \ref{tab:WordAndLineComparison}) demonstrate the clear superiority of line-level end-to-end OCR. This approach outperforms both word-level end-to-end pipelines and standard OCR systems. We benchmarked against five established word detection models (DBNet, DPText, TextFuseNet, CRAFT and MixNet) and six word recognition models (PARSeq, ABINet, MATRN, CCD, MAERec and SIGA). We also evaluated our pipeline against standard OCR systems, with results presented in Table \ref{tab:OCRSystemComparisonWeightedAvg}. All accuracies are calculated in a case-insensitive manner.
\subsection{Analysis}
\begin{figure*}[t]
    \centering
    \begin{center}
    $\begin{array}{@{\hspace{1pt}}c}
    \includegraphics[height=1.25in]{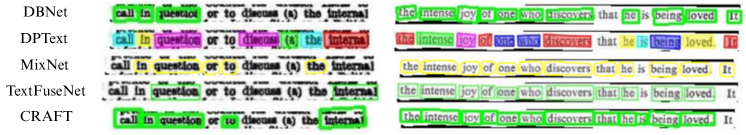}
    \end{array}$
    \end{center}
    \caption{Sample failure cases of word detection models.}
    \label{fig:failure_word}
\end{figure*}

The findings from our research reveal critical shortcomings within word detection models employed in end-to-end OCR systems. 
Some samples of such cases are shown in Fig.~\ref{fig:failure_word} and \ref{fig:failure_word_recognition}. \\
Moreover, our experiments highlight a significant oversight in word detection models regarding the recognition of punctuation and other textual elements within line-level images. This oversight cascades errors into subsequent recognition models, compounding inaccuracies throughout the OCR process. While recognition models themselves may demonstrate competence, the inherent inaccuracies in word detection represent a significant impediment, manifesting as a bottleneck in the overall document digitization process. \\
Additionally, our evaluation included an analysis of the efficiency of our proposed pipeline compared to traditional pipelines utilizing word detection and recognition models. Remarkably, our findings demonstrate that the line-level model significantly reduces processing time, making it a more efficient choice, particularly for edge devices. 
\begin{table}
\resizebox{1.0\linewidth}{!} 
{ 
\begin{tabular}{lcc}
	\toprule
	OCR System & CRR ($\uparrow$) & Flex Character Acc. ($\uparrow$) \\
	\midrule
	PP-OCR \cite{li2022ppocrv3} & 78.96 & 87.72 \\
	Tesseract \cite{smith2007overviewtesseract} & \textbf{88.16} & 88.41 \\
	DocTR \cite{doctr2021} & 83.52 & \color{blue}{97.14} \\
	\midrule
	Ours (Kraken + $\text{PARSeq}_\text{line}$) & \color{blue}{85.76} & \textbf{97.62} \\ 
	\bottomrule \\
\end{tabular}
}
\caption{Comparison of line-level ocr pipeline with some open-source frameworks on our {\em English} dataset}
\label{tab:OCRSystemComparisonWeightedAvg}
\end{table}
\begin{table}[t]
\begin{center}
\begin{tabular}{ lcc }
\toprule
OCR System & CRR ($\uparrow$) & Flex Character Acc. ($\uparrow$)
\\
\midrule
PP-OCR \cite{li2022ppocrv3} & 78.96 & 87.72 \\
Tesseract \cite{smith2007overviewtesseract} & 88.16 & 88.41 \\
DocTR \cite{doctr2021} & 83.52 & 97.14 \\
\midrule
Kraken + $\text{PARSeq}_\text{line}$ & 85.76 & 97.62 \\ 
Kraken + B.O + $\text{PARSeq}_\text{line}$ & {87.48} & 97.62\\
Kraken + T.O + $\text{PARSeq}_\text{line}$ & {89.74} & 97.62\\
Kraken + G.O + $\text{PARSeq}_\text{line}$ & {96.27} & 97.62\\
\bottomrule
\end{tabular}
\end{center}
\caption{Ablation study demonstrating the impact of mid-processing steps on CRR, using different line ordering strategies. B.O indicates blind ordering by line centroid, T.O and G.O represent ordering based on Tesseract predictions and Ground-truth, respectively, both employing Algorithm \ref{alg:algo-ordering}.}
\label{tab:OCRDrawbacks}
\end{table}

%% file: 6_ablations_discussion.tex
\section{Ablations and Discussion}
During our analysis of model accuracy, we encountered challenges associated with the Character Recognition Rate (CRR) metric. CRR evaluates accuracy based solely on the sequential ordering of text within an image, which may not adequately capture the nuances of document layout. To address this limitation, we explored alternative metrics and identified Flexible Character Accuracy (FCA) as a promising option. FCA considers the ordering of words at the line level, rather than enforcing a strict sequence of sentences, making it a more suitable metric for evaluating OCR accuracy. \\
Table \ref{tab:OCRSystemComparisonWeightedAvg} reveals the significant potential of our pipeline. Even without task-specific fine-tuning, it performs competitively with top OCR frameworks, suggesting even greater accuracy gains with further refinement. The observed drop in CRR accuracy of our proposed pipeline compared to Tesseract (Table \ref{tab:OCRSystemComparisonWeightedAvg}) highlights a specific issue with the line detection model used, particularly in correctly ordering sentences from left to right and top to bottom – the conventional reading direction for English documents. However, the higher FCA score indicates that if reading order is not a factor, our pipeline demonstrates superior effectiveness in document OCR. Consequently, our proposed pipeline represents a forward-looking approach to addressing OCR challenges, with a clear emphasis on achieving end-to-end accuracy.
\subsection{Different OCR Accuracy Metrics}
Character Recognition Rate (CRR) is a widely adopted metric, as discussed by Patel \etal \cite{patel2021feds}, embraced by many text recognition OCR models. It evaluates the accuracy of OCR systems by measuring the percentage of correctly identified characters compared to the ground truth text. The computation entails aligning recognized characters with their corresponding counterparts in the ground truth and calculating the ratio of correct identifications to the total number of characters. In our study, CRR served as one of the pivotal metrics in evaluating our pipelines.
\begin{equation}
	\text{CRR} = 1 - \frac{S + D + I}{S + D + C}.
\end{equation}
Here, $S$ denotes number of substitutions, $D$ denotes number of deletions, $I$ denotes number of insertions, $C$ denotes number of correct characters, and $N$ denotes number of characters in the reference, $N = S + D + C$. \\
Despite its widespread adoption, CRR suffers from several inherent limitations. One significant drawback lies in its heavy reliance on page and line segmentation methods, as well as the ordering of lines subsequent to the text line detection phase. Furthermore, CRR's inability to address contextual errors, such as misinterpretation of characters due to surrounding text or formatting inconsistencies, may hinder a comprehensive understanding of the model's end-to-end capabilities, leading to a significant decline in accuracy. \\
To address these shortcomings, we explored Flexible Character Accuracy (FCA), as introduced by Clausner \etal \cite{clausner2020flexible}. Unlike CRR, FCA is independent of line reading order, which is particularly advantageous given the challenge of predicting how a page of text will be read by humans, as well as the precise location of each character. By employing a hybrid computation that combines CRR at the line-level with the flexibility of page-level computations, FCA enables accurate measurement of order-independent accuracy, thereby enhancing the evaluation of OCR models.
\subsection{Limitations of CRR: An Ablation Study}
As previously noted, Character Recognition Rate (CRR) may not be the optimal metric for evaluating the accuracy of end-to-end Optical Character Recognition (OCR) pipelines. Table \ref{tab:OCRDrawbacks} presents an ablation study designed to highlight the potential for CRR to fluctuate and offer misleading representations of overall OCR model performance. This demonstration was achieved by incorporating various pre-processing steps prior to line recognition. While these additional steps were included for illustrative purposes, they are not considered practical for real-world implementation and have therefore been excluded from the original table. \\
CRR fails to capture the holistic accuracy of the system in conveying the intended meaning to the reader. In standard reading practices, English text is typically read from left to right. However, in document images, the order in which lines of text are read is often unpredictable due to diverse layouts and individual preferences. When evaluating end-to-end OCR systems, the primary goal is to assess the system's overall accuracy. \\
However, using CRR for this purpose can be misleading. CRR aggregates all lines of text into a single sequence, comparing it to the ground truth. This approach means that even a single incorrect or missing line of text can significantly impact the CRR of other, correctly predicted lines. \\
To address this, we utilized FCA \cite{clausner2020flexible} metric specifically for evaluating end-to-end OCR accuracy. This metric prioritizes the overall performance of the OCR pipeline, rather than being overly sensitive to the specific order in which text is recognized. This approach provides a more accurate and stable assessment of an OCR system's true capability, as it mitigates the misleading effects of CRR when evaluating complex document layouts.
\begin{algorithm}[H]
  \caption{Kraken Lines Reordering}\label{alg:algo-ordering}
  \begin{algorithmic}[1]
    \Require Similarity threshold $\tau$ (e.g., $90\%$)
    \ForAll{images}
      \State $R \gets [R_1,\dots,R_n]$\Comment{reference model line predictions}
      \State $K \gets [K_1,\dots,K_n]$\Comment{Kraken+PARSeq$_\text{line}$ predictions}
      \ForAll{lines $R_i$ in $R$}
        \State $j \gets \arg\min_j\,D(R_i, K_j)$
        \State $S \gets 1 - \dfrac{D(R_i, K_j)}{\max\bigl(|R_i|,|K_j|\bigr)}$
        \If{$S \ge \tau$}
          \State remove $K_j$ from $K$ \Comment{store match}
          \State insert $K_j$ at position $i$ in $K$
        \EndIf
      \EndFor
    \EndFor
  \end{algorithmic}
\end{algorithm}
\subsection{Kraken for Text Detection}
Kraken \cite{kiessling2020modularkraken} utilizes a machine learning-based approach for text line detection. The primary motivation for utilizing Kraken for line segmentation stems from its integrated approach, combining layout analysis with text line detection. This streamlined process eliminates the need for separate layout analysis tools, thereby reducing the complexity and computational overhead of the overall document processing pipeline. The model consists of the following key stages:
\begin{enumerate}
    \item \textbf{Preprocessing} Images are normalized for consistent analysis, typically involving noise reduction and binarization.
    \item \textbf{Baseline Detection} Kraken employs a neural network model to predict the baseline for each text line. This model has been trained on a variety of scripts and document styles.
    \item \textbf{Region of Interest (ROI) Generation} Based on the detected baselines, rectangular ROIs are created, encompassing the estimated text lines.
    \item \textbf{Line Segmentation Refinement} Post-processing techniques are applied to refine the line segmentation, ensuring accurate separation of lines even in cases of overlapping or touching text.
\end{enumerate}
The performance of Kraken was evaluated against other established line-detection models using their respective line-detection checkpoints. As demonstrated in Table \ref{tab:KrakenComparison}, Kraken consistently outperformed the alternative methods, thus justifying its selection for inclusion in our proposed pipeline.
\begin{table}[t]
\begin{center}
\begin{tabular}{ lcc }
\toprule
Line Detection & Line Recognition & CRR ($\uparrow$) \\
\midrule
DBNet \cite{liao2020Dbnet} & $\text{PARSeq}_\text{line}$ & 24.30 \\
TFN \cite{ye2020textfusenet} & $\text{PARSeq}_\text{line}$ & 28.10 \\
DPText \cite{ye2023dptextdetr} & $\text{PARSeq}_\text{line}$ & 28.44 \\
SRFormer \cite{bu2023srformer} & $\text{PARSeq}_\text{line}$ & 29.19 \\
Doc-UFCN \cite{boillet2021multiple} & $\text{PARSeq}_\text{line}$ & 53.97 \\
Mixnet \cite{zeng2023mixnet} & $\text{PARSeq}_\text{line}$ & 78.58 \\
\midrule
Tesseract \cite{smith2007overviewtesseract} & $\text{PARSeq}_\text{line}$ & 80.79 \\
DocTR \cite{doctr2021} & $\text{PARSeq}_\text{line}$ & 82.43 \\
\midrule
Kraken \cite{kiessling2020modularkraken} & $\text{PARSeq}_\text{line}$ & 85.76 \\ 
\bottomrule
\end{tabular}
\end{center}
\caption{Performance comparison of Kraken against alternative line detection models.}
\label{tab:KrakenComparison}
\end{table}

%% file: 7_conclusion.tex
\section{Conclusion}
This work establishes a foundation for line-level OCR research, transcending conventional word-level pipelines. The curated English page dataset is a valuable resource for developing end-to-end OCR solutions. Our analysis highlights the potential of line-level OCR within broader pipelines of Document OCR.  Additionally, we demonstrate significant efficiency gains when employing line-level OCR. The results demonstrate the need for further refinement of line-level OCR methods, opening avenues for continued advancements. \\
Moreover, our end‑to‑end experiments demonstrate that line‑level OCR not only improves raw recognition accuracy, but also substantially reduces cascading errors caused by wrongly segmented or wrongly ordered words. By processing entire lines as coherent units, the model leverages broader contextual cues like sentence‑level syntax and punctuation patterns, which leads to more robust handling of distortions and ambiguous characters. \\
The creation of our 251‑page, line‑annotated dataset fills a critical gap in publicly available benchmarks. We believe this resource will catalyze further innovations in document OCR, particularly in integrating powerful language models to correct residual errors. Collectively, these contributions position line‑level OCR as a promising direction for next‑generation, high‑accuracy document understanding systems.

%% file: 8_future_work.tex
\section{Future Work}
Transitioning to line-level OCR not only streamlines document analysis by eliminating word detection but also creates a powerful synergy with language models. The richer contextual information within line-level text significantly improves the overall accuracy and allows for integration of language models (LMs) in the proposed OCR pipeline in future. \\
The field's evolution, alongside advancements in GPU capabilities, compels research to move beyond a focus on isolated stages. Embracing line-level OCR not only offers a promising avenue for enhancing real-world document understanding but also streamlines the OCR pipeline. Our experiments demonstrate the potential to eliminate a separate word detection model, facilitating efficient deployment on edge devices.

%% file: appendix.tex
\section{Supplementary Material}
\subsection{Deep Dive into TRDG Data Generation}
For line-level images, we utilize the Text Recognition Data Generator (TRDG\footnote{\url{https://github.com/Belval/TextRecognitionDataGenerator}}) for synthetic images generation. We've curated a diverse font collection of $3309$ English fonts to ensure realistic text rendering. Our English corpus comprises $1.21$ million text lines containing $600337$ distinct words sourced from Wikipedia.
Along with standard characters, we tried including commonly used special characters and numerals into our vocabulary to enhance its robustness. TRDG's flexibility enables us to simulate a variety of image distortions during synthetic image generation. Specifically, we have included the following distortions: 
\begin{enumerate}
	\item \textbf{Blurring:} Text undergoes blurring operations with varying intensity. Simulates real-world image degradation, forcing the model to learn robust features even under less-than-ideal conditions.
	\item \textbf{Background Variation:} Images are superimposed on diverse backgrounds, including white, coloured, ruled-line, figures, reversed-blurred text, and quasicrystals. Text colour is also varied. This allows the model to differentiate text from complex and potentially interfering backgrounds.
	\item \textbf{Noise:} We introduce different noise types, such as Gaussian, salt and pepper, dilation and erosion, on synthetic images at random. These capabilities are integrated with TRDG's existing image generation code to further increase the diversity of generated images. It reduces the reliance on pristine images, improving the model's resilience to common artefacts that affect scanned documents.
	\item \textbf{Skewing and shearing:} Text images are skewed and sheared within a range of -10 to +10 degrees. The model must adapt to text that may not be perfectly aligned, a frequent occurrence in real-world scenarios.
	\item \textbf{Spacing:} Margins and inter-character spacing are randomly altered. This allows the model to handle non-uniform, irregular text layouts often seen in handwritten or older documents.
\end{enumerate}
\subsection{Additional Dataset Images}
\begin{figure*}
    \centering
    \begin{center}
    $\begin{array}{@{\hspace{5pt}}c@{\hspace{5pt}}c@{\hspace{5pt}}c}
    \includegraphics[height=3.0in]{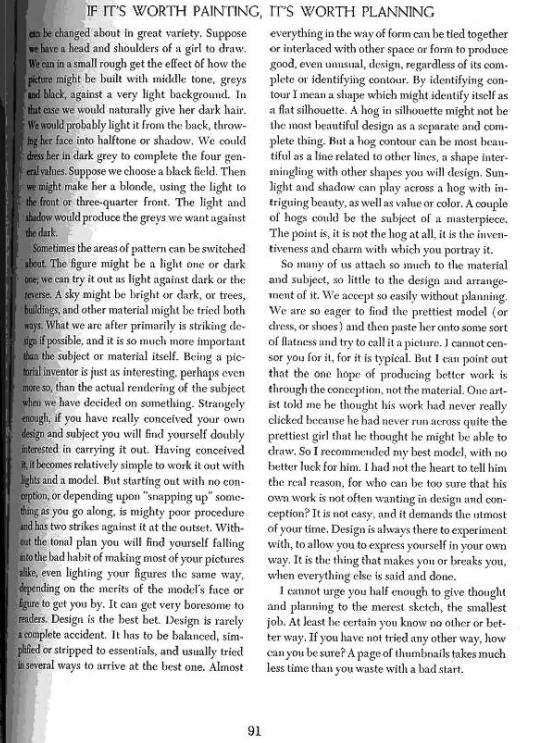}&
    \includegraphics[height=3.0in]{IM_6.jpg}\\
    \end{array}$
    $\begin{array}{@{\hspace{5pt}}c@{\hspace{5pt}}c@{\hspace{5pt}}c}
    \includegraphics[height=2.0in]{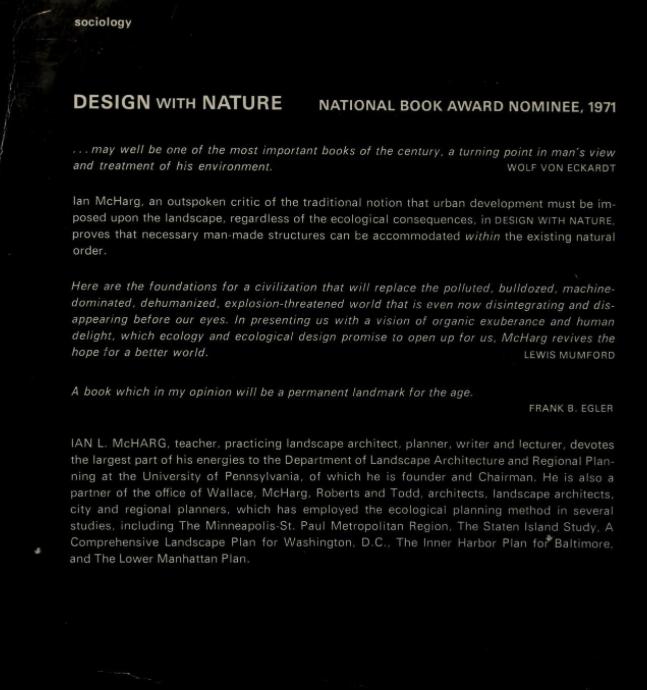}&
    \includegraphics[height=2.0in]{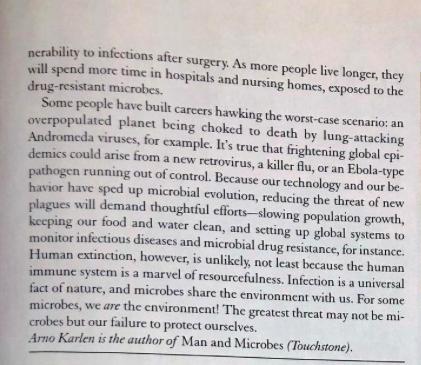}\\
    \end{array}$
    $\begin{array}
    {@{\hspace{5pt}}c@{\hspace{5pt}}c@{\hspace{5pt}}c}
    \includegraphics[height=3.0in]{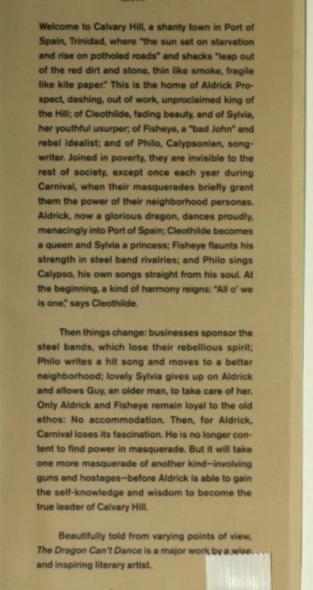}&
    \includegraphics[height=3.0in]{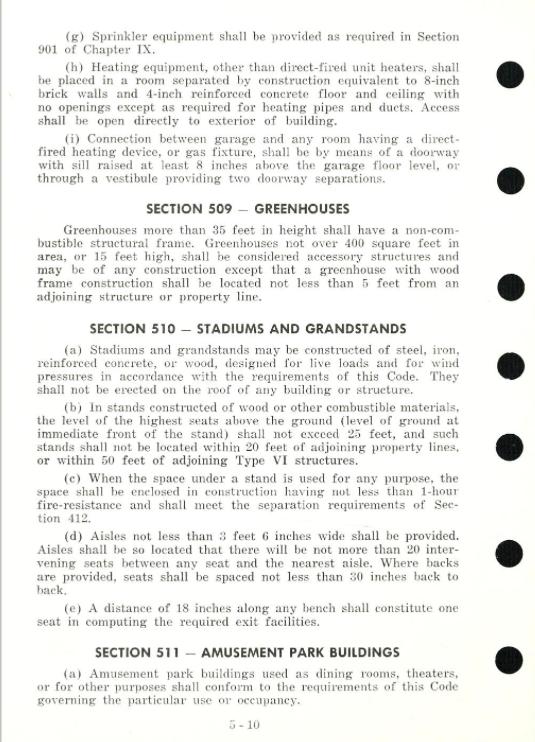}&
    \includegraphics[height=3.0in]{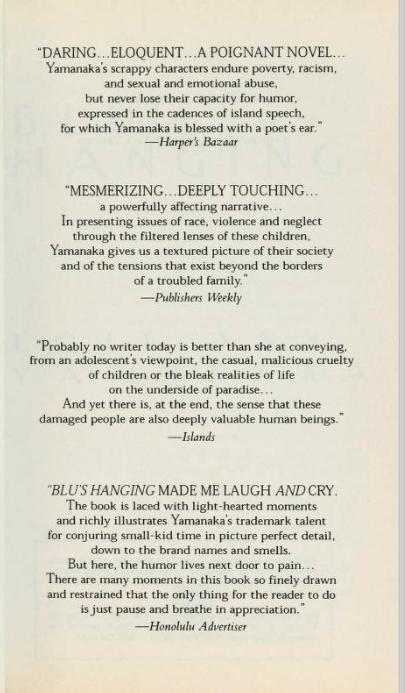}\\
    \end{array}$
    \end{center}
    \caption{Additional sample images from the proposed dataset}
    \label{fig:suppl_dataset}
\end{figure*}

Figure \ref{fig:suppl_dataset} presents additional sample images from the proposed dataset, which encompasses a diverse range of fonts, backgrounds, designs, and other variations. Our experimentation with multi-column text images underscores the need for detection models to prioritize both layout understanding and end-to-end efficiency while maintaining high accuracy.
\subsection{\textbf{PARSeq for Text Recognition at Line-level}}
PARSeq (Permuted Autoregressive Sequence Model) has emerged as a leading model in Scene Text Recognition (STR), a critical component of Optical Character Recognition (OCR) systems. Its unique architecture and training methodology offer several key advantages that contribute to its superior performance:
\begin{enumerate}
    \item \textbf{Unified Architecture} PARSeq integrates context-free non-autoregressive and context-aware autoregressive inference within a single model. This unification allows for efficient and effective decoding, leveraging both global and local image features.
    \item \textbf{Permutation Language Modeling} PARSeq employs Permutation Language Modeling (PLM) during training. PLM encourages the model to learn an ensemble of internal language models with shared weights, enhancing robustness to noise and variations in text appearance.
    \item \textbf{Bidirectional Context} PARSeq utilizes bidirectional context during inference, allowing the model to incorporate information from both preceding and following characters. This bidirectional attention mechanism improves accuracy, especially in challenging cases like degraded or partially occluded text.
    \item \textbf{Parallel Token Processing} PARSeq processes tokens in parallel, enabling efficient computation and faster inference compared to traditional sequential models.
    \item \textbf{Robustness to Diverse Text} PARSeq demonstrates exceptional robustness to various text characteristics, including different fonts, styles, orientations, and languages. This adaptability makes it well-suited for real-world OCR applications with diverse text inputs.
    \item \textbf{Iterative Refinement} A notable feature of PARSeq is its ability to perform iterative refinement. This process starts with an initial prediction. The output from this initial prediction is then used as context for the next iteration, leveraging bidirectional attention (cloze context) to refine the results. This iterative approach allows PARSeq to progressively improve its predictions, incorporating information from both preceding and following characters, leading to enhanced accuracy and robustness in challenging scenarios. This iterative refinement can be applied multiple times, offering flexibility to balance accuracy and computational cost based on specific requirements.
    \item \textbf{State-of-the-art Performance} PARSeq has consistently achieved state-of-the-art results on several benchmark datasets, outperforming previous models in terms of accuracy, efficiency, and computational cost.
\end{enumerate}
